\documentclass[nohyperref]{article}

\pdfoutput=1

\usepackage{microtype}
\usepackage{graphicx}
\usepackage{subfigure}
\usepackage{booktabs} 

\usepackage{hyperref}

 \usepackage[accepted]{icml2022_TAGML}

\usepackage{amsmath}
\usepackage{amssymb}
\usepackage{mathtools}
\usepackage{amsthm}

\usepackage[capitalize,noabbrev]{cleveref}

\usepackage{soul}

\theoremstyle{plain}
\newtheorem{theorem}{Theorem}[section]

\newtheorem{corollary}[theorem]{Corollary}
\theoremstyle{definition}
\newtheorem{definition}[theorem]{Definition}

\theoremstyle{remark}

\usepackage[textsize=tiny]{todonotes}

\icmltitlerunning{Zeroth-Order Topological Insights into Iterative Magnitude Pruning}

\begin{document}

\twocolumn[
\icmltitle{Zeroth-Order Topological Insights into Iterative Magnitude Pruning}




\begin{icmlauthorlist}
\icmlauthor{Aishwarya Balwani}{sch}
\icmlauthor{Jakob Krzyston}{sch}
\end{icmlauthorlist}

\icmlaffiliation{sch}{School of Electrical \& Computer Engineering, Georgia Institute of Technology, Atlanta, GA, USA}

\icmlcorrespondingauthor{Aishwarya Balwani}{abalwani6@gatech.edu}
\icmlcorrespondingauthor{Jakob Krzyston}{jakobk@gatech.edu}

\icmlkeywords{Pruning, Persistent Homology, Topology, Neural Networks}

\vskip 0.3in
]



\printAffiliationsAndNotice{}

\begin{abstract}
Modern-day neural networks are famously large, yet also highly redundant and compressible;
there exist numerous pruning strategies in the deep learning literature that yield over $90\%$ sparser sub-networks of fully-trained, dense architectures while still maintaining their original accuracies.
Amongst these many methods though -- thanks to its conceptual simplicity, ease of implementation, and efficacy -- Iterative Magnitude Pruning (IMP) dominates in practice and is the de facto baseline to beat in the pruning community.
However, theoretical explanations as to why a simplistic method such as IMP works at all are few and limited.
In this work, we leverage the notion of persistent homology to gain insights into the workings of IMP and show that it inherently encourages retention of those weights which preserve topological information in a trained network.
Subsequently, we also provide bounds on how much different networks can be pruned while perfectly preserving their zeroth order topological features, and present a modified version of IMP to do the same.
\end{abstract}

\section{Introduction}
\label{intro}
The many successes of deep neural networks (DNNs) across domains such as computer vision \cite{simonyan2014very,he2016deep}, speech recognition \cite{graves2013speech}, natural language processing \cite{vaswani2017attention, brown2020language}, biomedicine \cite{ronneberger2015u, rajpurkar2017chexnet}, and bioinformatics \cite{jumper2021highly} have made them ubiquitous in both academic and industrial settings.
However, while DNNs boast of being able to achieve state of the art results on a plethora of complex problems, they also have the dubious honour of being untenably large and unsuitable for applications with power, memory, and latency constraints.
One way of tackling these issues is to reduce the parameter-counts of DNNs, thereby decreasing their size and energy consumption while improving inference speeds.
As a result, the field of \emph{neural network pruning} which studies techniques for eliminating unnecessary weights in both pre-trained and randomly initialized DNNs without loss of accuracies \cite{mozer1988skeletonization, hanson1988comparing,lecun1989optimal, hassibi1992second} has seen renewed interest in recent years \cite{han2015learning,li2016pruning, cheng2017survey, frankle2018lottery, lee2018snip, blalock2020state}.

\subsection{Related Work}
Increased activity on the methodological front has subsequently spurred principled analytical efforts to help explain when or how various pruning methods ostensibly work.

For instance, by way of deriving generalization bounds for DNNs via compression \cite{arora2018stronger}, previous work has provided some theoretical justification for pruned sub-networks.
Iterative Magnitude Pruning (IMP) has been explored via the observation that sparse sub-networks that maintain accuracies of the original network are stable to stochastic gradient descent noise and optimize to linearly connected minima in the loss landscape \cite{frankle2020linear}.
A related empirical work has also looked into how fundamental phenomena such as  weight evolution and emergence of distinctive connectivity patterns are affected by changes in the iterative pruning procedure \cite{paganini2020iterative}.
A gradient-flow based framework \cite{lubana2020gradient} has been used to show why certain importance measures work for pruning early on in the training cycle.

More recently, a number of works have also begun studying pruning at initialization, providing insights into gradient-based pruning via conservation laws \cite{wang2020picking, tanaka2020pruning}, presenting theoretical analyses of pruning schemes that fall under the purview of sensitivity-based pruning \cite{hayou2020robust}, developing a path-centric framework for studying pruning approaches \cite{gebhart2021unified}, and looking at magnitude pruning in linear models trained using gradient flow \cite{elesedy2020lottery}.

Unfortunately, despite the flurry of contemporary work and results in the area, a mathematically rigorous yet intuitive explanation for \emph{why} IMP works well remains missing.

\subsection{Contributions}
Given its integral place in the present DNN pruning research landscape, there is a strong impetus to establish a precise but flexible framework which uses the same language to speak of not only IMP, but also related problems of theoretical and empirical interest such as the Lottery Ticket Hypothesis (LTH) \cite{frankle2018lottery}, DNN initialization and generalization \cite{morcos2019one}, weight rewinding, and fine-tuning \cite{renda2020comparing}.

Towards this end we utilize the formalism of algebraic topology, which has found increasing application in the characterization of DNN properties such as learning capacity \cite{guss2018characterizing}, latent and activation space structure \cite{khrulkov2018geometry, gebhart2019characterizing, carlsson2020topological}, decision boundaries \cite{ramamurthy2019topological}, and prediction confidence  \cite{lacombe2021topological}.
Specifically, we use \emph{neural persistence} \cite{rieck2018neural} -- a measure based on persistent homology for assessing the topological complexity of neural networks -- to ascertain which set of weights in the DNN capture its zeroth-dimensional topological features, and show that IMP with high probability preserves them. 
Following this result, the main contributions of our work are:
\begin{itemize}
    \vspace{-1mm}
    \item A formal yet intuitive framework rooted in persistent homology that can reason about magnitude-based pruning at large and other phenomena related to it.
    \vspace{-1mm}
    \item A mathematically-grounded perspective on IMP that helps explain its empirical success through supporting theoretical lower bounds regarding its ability to preserve topological information in a trained DNN. 
    \vspace{-1mm}
    \item Precise upper bounds on the maximum achievable compression ratios for fully-connected, convolutional, and recurrent layers such that they maintain their zeroth-dimensional topological features, and realizations of the same for some established architecture-dataset pairings in the pruning literature.
    \vspace{-1mm}
    \item A topologically-driven algorithm for iterative pruning, which would perfectly preserve their zeroth-order topological features throughout the pruning process.
\end{itemize}

\section{Background \& Notation}
\label{background}

In this section we offer some background on neural network pruning, persistent homology, and neural persistence, while also establishing the relevant notation.

\subsection{Iterative Magnitude Pruning}
\label{sec:imp_bg}

A neural network \emph{architecture} is a function family $f(x;\cdot)$, where the architecture consists of the configuration of
the network’s parameters and the sets of operations it uses to produce outputs from inputs, including the arrangement
of parameters into convolutions, activation functions, pooling, batch norm, etc.
A \emph{model} is a particular instantiation of an architecture, i.e., $f(x; \mathcal{W})$ with specific parameters $\mathcal{W}$.

Neural network \emph{pruning} entails taking as input a model $f(x; \mathcal{W})$ and producing a new model $f(x; M \odot \mathcal{W}^*)$ where $M \in \{0,1\}^{|\mathcal{W}^*|}$ is a binary mask that fixes certain parameters to 0, $\odot$ is the elementwise product operator, and $\mathcal{W}^*$ is a set of parameters that may differ from $\mathcal{W}$.
A number of different heuristics called \emph{scoring functions} may be used to construct the mask $M$, which decide which weights are to be pruned or kept.
Popular scoring functions include the magnitude of the weights or some form of the gradients of a specified loss with respect to the weights.
If the mask $M$ is constructed by scoring the parameters of a model per layer, the pruning scheme is said to be \emph{local}, whereas if $M$ is constructed by scoring all parameters in the set $\mathcal{W}$ collectively, the pruning scheme is said to be \emph{global}. 

The \emph{sparsity} of the {pruned} model is $\frac{|f(x;M \odot \mathcal{W}^*)|_{\textrm{nnz}}}{|\mathcal{W}|}$ where  $|\cdot|_{\textrm{nnz}}$ is a function that counts the number of non-zeros of the pruned model and $|\mathcal{W}|$ is the total number of parameters in the original model. 
The \emph{compression ratio} ($\eta$) is given as $\frac{|\mathcal{W}|}{|f(x;M \odot \mathcal{W}^*)|_{\textrm{nnz}}}$ which is simply the inverse of the sparsity.

Given an initial untrained model $f(x,\mathcal{W}_0)$, \emph{iterative magnitude pruning} (IMP) takes the following steps to obtain a sparsified model $f(x;\mathcal{W}_N)$ with $p\%$ target sparsity:
\begin{enumerate}
    \item Train $f(x;\mathcal{W}_0)$ for $t$ iterations, thereby obtaining the intermediate parameters $\mathcal{W}_{0,t}$
    \item Mask the non-zero $\frac{p}{N}$\% parameters of lowest magnitude in $\mathcal{W}_{0,t}$ to arrive at parameters $\mathcal{W}_1$.
    \item Repeat the aforementioned steps $N$ times.
    
\end{enumerate}

\subsection{Persistent Homology}

Persistent homology \cite{edelsbrunner2008persistent} is a tool commonly used in topological data analysis (TDA) to understand high-dimensional manifolds, and has been successfully employed in a range of applications such as analysing natural images \cite{carlsson2008local}, characterizing graphs \cite{sizemore2017classification, rieck2017clique}, and finding relevant features in unstructured data \cite{lum2013extracting}.

First, however, the space of interest must be represented as a \emph{simplicial complex} which is effectively the extension of the idea of a graph to arbitrarily high dimensions.
The sequence of \emph{homology groups} \cite{edelsbrunner2022computational} of the simplicial complex then formalizes the notion of the topological features which correspond to the arbitrary dimensional ``holes" in the space.
For example, holes of dimension 0, 1, and 2 refer to connected components, tunnels, and voids respectively in the space of interest.
Information from the $d^{\textrm{th}}$
homology group is summarized by the $d^{\textrm{th}}$ \emph{Betti number} ($\beta_d$) which
merely counts the number of $d$-dimensional holes;
thus a circle has Betti numbers (1, 1),
i.e., one connected component and one tunnel, while a disc has Betti numbers (1, 0), i.e., one
connected component but no tunnel.

Betti numbers themselves unfortunately are of limited use in practical applications due to their instability and extremely coarse nature, which has prompted the development of {persistent homology}.
Given a simplicial complex $K$ with an additional set of
scales $a_0 \leq a_1 \leq ... \leq a_{m-1} \leq a_m$, one can put $K$ through a filtration, i.e., a nested sequence of simplicial complices $\emptyset = K_0 \subseteq K_1 \subseteq ... K_{m-1} \subseteq K_m = K$.
The filtration essentially represents the growth of $K$
as the scale is changed, and during this process topological  features can be created (new
vertices may be added, for example, which creates a new connected component) or destroyed (two
connected components may merge into one).

\emph{Persistent homology} tracks these changes, and represents
the creation and destruction of a feature as a point $(a_i,a_j)\in \mathbb{R}^2$ for indices $i \leq j$ with respect to
the filtration.
The collection of all points corresponding to $d$-dimensional topological features is called the $d^{\textrm{th}}$ \emph{persistence diagram} ($\mathcal{D}_d$),
and can be thought of as a collection of Betti numbers at multiple
scales.
Given a point $(x,y) \in \mathcal{D}_d$, the quantity $\operatorname{pers}(x, y) := |y - x|$ is referred to as its \emph{persistence}, where $|\cdot|$ is an appropriate metric.
Typically, high persistence is considered to correspond to features, while low persistence is considered
to indicate noise \cite{edelsbrunner2000topological}.

\subsection{Neural Persistence}

\begin{figure}[ht]
\vskip 0.1in
\begin{center}
\centerline{\includegraphics[width=\columnwidth]{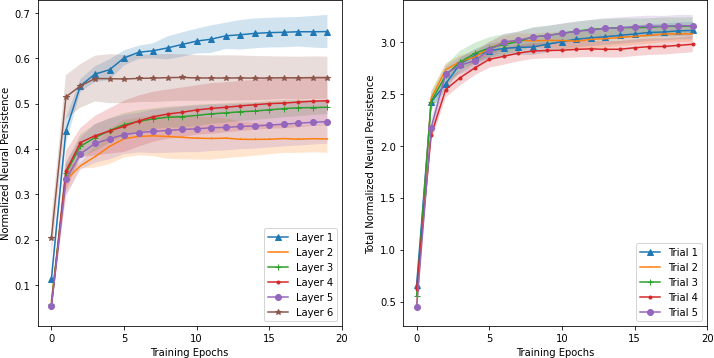}}
\caption{For a fully-connected, 6-layer network trained on the MNIST dataset across 20 epochs of training, (Left) Layer-wise normalized neural persistences and (Right) Trial-wise total neural persistences for the entire network. Means are presented as solid lines, standard deviations are shaded. }
\label{fig/NP_plots_dense}
\end{center}
\vskip -0.3in
\end{figure}

{Neural persistence} is a recently proposed measure of structural complexity for DNNs \cite{rieck2018neural} that exploits both network architecture and weight information through persistent homology, to capture how well trained a DNN is.
For example, one can empirically verify that the ``complexity" of a simple, fully connected network as measured by neural persistence increases with learning (Fig. \ref{fig/NP_plots_dense}).

\begin{figure}[ht]
\begin{center}
\centerline{\includegraphics[width=\columnwidth]{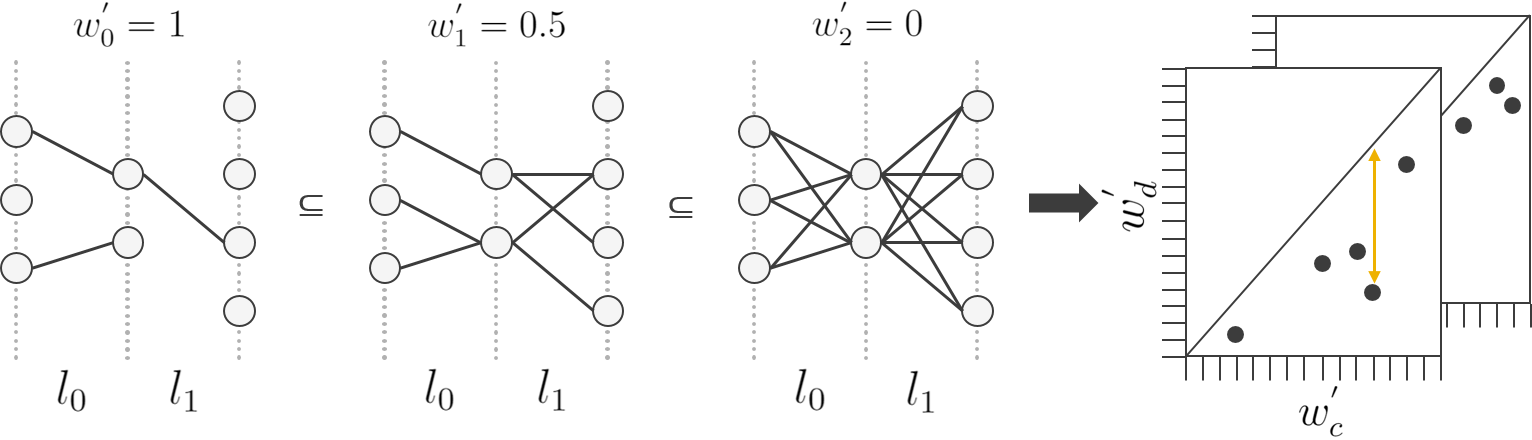}}
\end{center}
\vskip -0.2in
\caption{Persistence diagrams show how long a particular model ``persists" as the network undergoes the defined filtration (i.e, pruning).
From left to right, as the weight threshold, $w^{'}$, decreases, and the number of connections kept in the two layers ($l_{0}$ and $l_{1}$) increases. On the right, the persistence diagram plots the duration of each structure as a coordinate corresponding to the weight threshold at which the structure was created $(w^{'}_{c})$, and the weight threshold at which it was destroyed $(w^{'}_{d})$.
The most prominent structure has the greatest \textit{persistence}, which is measured by its distance from the diagonal.
In this example, the point indicated by the yellow arrow would indicate the most persistent feature.}
\label{fig:persistent}
\vskip -0.1in
\end{figure}

Construction of the measure itself relies on the idea that one can view a model $f(x;\mathcal{W})$ as a stratified graph $G$ with vertices $V$, edges $E$, with a mapping function $\varphi: E \rightarrow \mathcal{W}$ that allows for the calculation of the persistent homology of \emph{every layer} $G_k$ in the model using a filtration induced by sorting the weights.
More precisely, given its set of weights $\mathcal{W}_k$ at any training step, let $w_{\max }:=\max _{w \in \mathcal{W}_k}|w|$, and $\mathcal{W}_k^{\prime}:=\left\{|w| / w_{\max } \mid w \in \mathcal{W}\right\}$ be the set of transformed weights indexed in non-ascending order, such that $1=w_{0}^{\prime} \geq w_{1}^{\prime} \geq ... \geq 0$. This permits one to define a filtration for the $k^{\textrm{th}}$ layer $G_{k}$ as $G_{k}^{(0)} \subseteq G_{k}^{(1)} \subseteq \ldots$, where $G_{k}^{(i)}:=\left(V_{k} \sqcup V_{k+1},\left\{(u, v) \mid(u, v) \in E_{k} \wedge \varphi^{\prime}(u, v) \geq w_{i}^{\prime}\right\}\right)$, and
$\varphi_{k}^{\prime}(u, v) \in \mathcal{W}_{k}^{\prime}$ denotes the transformed weight of an edge.
The relative strength of a connection is thus preserved by the filtration, and weaker weights with $|w| \approx 0$ remain close to 0.
Additionally, since $w^{\prime} \in[0,1]$ for the transformed weights, the filtration makes the network invariant to scaling of $\mathcal{W}$, simplifying the comparison of different networks.
Using this filtration one can calculate the persistent homology for every layer $G_k$.
As the filtration contains at most 1-simplices (edges), the topological information captured is zero-dimensional, i.e. reflects how connected components are created and merged during the filtration, and can be shown graphically with a $0^{\textrm{th}}$ persistence diagram (Fig. \ref{fig:persistent}).

\emph{Neural persistence} of the $k^{\textrm{th}}$ layer $G_k$ of a DNN is then defined as the $p$-norm of the persistence diagram $\mathcal{D}_k$ resulting from the previously discussed filtration, i.e.,
\begin{equation*}
\setlength{\abovedisplayskip}{2pt}
\setlength{\belowdisplayskip}{2pt}
\textrm{NP}(G_k) := ||\mathcal{D}_k||_p := \left(\sum_{(c, d) \in \mathcal{D}_{k}} \operatorname{pers}(c, d)^{p}\right)^{\frac{1}{p}}  
\end{equation*}

Typically $p = 2$, which captures the Euclidean distance of the points in $\mathcal{D}_k$ to the diagonal.

We note that the above definition strictly corresponds to fully-connected layers, but the notion can easily be extended to convolutional and recurrent layers by representing the former using appropriately sized Toeplitz matrices \cite{goodfellow2016deep} and the latter as multiple ``unrolled" fully-connected layers, with the same, shared set of weights.

Neural persistence can also be normalized to values in $[0,1]$ in a scale-free manner, thus providing a simple method to compare the structural complexities of differently sized layers\footnote{For reference see Theorem 1 \cite{rieck2018neural}.} across various architectures.
The total neural persistence of a model with $L$ layers is given by the sum of all the individual layerwise neural persistences, i.e.,

\vskip -0.1in
\begin{equation*}
\setlength{\abovedisplayskip}{2pt}
\setlength{\belowdisplayskip}{2pt}
    \operatorname{NP}(G) := \sum_{k = 1}^{L} \operatorname{NP}(G_k)  
\end{equation*}

\section{A Topological Perspective on Magnitude-Based Pruning}
\label{theory}

This section details our interpretation of magnitude pruning (MP) via the lens of neural persistence, followed by some insights we glean regarding IMP from this novel perspective.
Consequently, we provide a lower bound on the relative topological information that is retained by IMP at every iteration, define a quantity that gives an upper bound on how much different types of neural network architectures may be pruned while still conserving its zeroth-dimensional topological features, and present a topologically-motivated version of IMP that guarantees the same.

\subsection{Neural Persistence \& Magnitude Pruning}

Here we explicitly mention the key aspects of neural persistence which can be deduced from the background covered in Section \ref{background} to arrive at a topological understanding of MP:

\begin{itemize}
    \vspace{-1mm}
    \item Neural persistence relies on a super-level set filtration \cite{cohen2009extending, bubenik2015statistical} and sorts only the edges of $G_k$, the DNN layer\footnote{ Every $G_k$ is a bipartite graph.} it acts on.
    \vspace{-1mm}
    \item The weights $\mathcal{W}_k^{\prime}$ of a layer $G_k$ are normalized to values in $[0,1]$, disregarding the signs of the weights $\mathcal{W}_k$ while still respecting their relative magnitudes.
    \vspace{-1mm}
    \item All the vertices of $G_k$ are already present at the beginning of the filtration and result in $m_k + n_k$ connected components at the start of the filtration, where $m_k, n_k$ are the cardinalities of the two vertex sets of $G_k$.
    \vspace{-1mm}
    \item Entries in the corresponding zeroth-dimensional persistence diagram $\mathcal{D}_k$ are of the form $(1,x), x \in \mathcal{W}_k^{\prime}$, and are situated below the diagonal.
    
    \item As the filtration progresses, the weights greater than the threshold $a_i$ are introduced in the zeroth-order persistence diagram, so long as it connects two vertices in $G_k$ without creating any cycles \cite{lacombe2021topological}.
    
    \item As a result, the filtration ends up with the maximum spanning tree\footnote{For a proof, see Lemmas 1 \& 2 in \citeauthor{doraiswamy2020topomap}}$^{,}$\footnote{For a visual explanation of the filtration, see Appendix \ref{appx/superlevel_filtration}.} (MST) of the graph $G_k$.

\end{itemize}
\vskip -0.1in

From the \ul{viewpoint of persistent homology} this implies that all the zeroth-order topological information of $G_k$ as captured by its neural persistence is encapsulated in the weights of $\mathcal{W}_k^{\prime}$ which form its MST.
    
From the \ul{standpoint of magnitude-based pruning}\footnote{Since neural persistence by its current definition is applied layer-wise, the insights we gain from using it through the rest of this paper correspond to local pruning.} we have a novel topologically-motivated scoring function, whose goal is to maintain the zeroth-order topological information\footnote{While it is true that the notion of  persistent homology (and ergo neural persistence) on a set of weights $\mathcal{W}_k^{\prime}$ could be extended beyond the zeroth dimension, thus implying higher orders of topological information that are not captured by the current measure, we note that previous work has found that zero-dimensional
topological information still captures a significant portion of it \cite{rieck2016exploring, hofer2017deep}, thereby sufficing for now.} in a set of weights, and the resulting mask $M$ prunes any weights that are not part of the MST of a particular layer.

Consequently, we arrive at the following insight regarding IMP and its practical efficacy:

\vskip 0.1in
\noindent
\framebox{\parbox{\dimexpr\linewidth-2\fboxsep-2\fboxrule}{\itshape
  At every iteration, the weights retained by IMP in a layer are likely to overlap significantly with those present in its MST with relatively high probability.
  This ensures the pruning step itself does not severely degrade the zeroth-order topological information learnt by the layer in the previous training cycle, and in turn provides the network with a sufficiently informative initialization, allowing it train to high levels of accuracy once again.}}
 \vskip 0.1in

In the following subsections, we further formalize this intuition by deriving a definitive lower bound on the expected overlap between the weights in a layer's MST and those retained by IMP, as well as a strict upper bound on how much a layer can be pruned while still maintaining its zeroth-dimensional topological information.

\subsection{Topologically Critical Compression Ratio}

Building off the observation that we only need as many weights as that of the MST of a layer $G_k$ to maintain its zeroth-dimensional topological features, we define the following quantity that allows us to achieve maximal topologically conservative compression.

\begin{definition}
\label{def:topological_compression_ratio}
The topologically critical compression ratio ($\eta_{\tau}$) for any graph $G$ with edges $E$, vertices $V$, and mapping function $\varphi: E \rightarrow \mathcal{W}$ is defined as the quantity
$$\eta_{\tau} := \frac{|\mathcal{W}|}{|\operatorname{MST}(G)|}$$
\end{definition}
where the function $\operatorname{MST(\cdot)}$ denotes the MST of the graph in question and $|\cdot|$ is the cardinality of a set.

$\eta_{\tau}$ is the maximal achievable compression for the graph $G$ which would still be able to perfectly preserve the zeroth-order topological complexity of $G$, assuming the right set of weights (i.e., those in the MST of $G$) are retained.   

Using Def.~\ref{def:topological_compression_ratio}, in the context of DNN pruning we subsequently arrive at the following result

\begin{theorem}
\label{thm:top_compression_ratios}
Given a layer $G_k$ with weights $\mathcal{W}_k^{\prime}$ joining $m_k$ input nodes to $n_k$ output nodes, for any compression ratio $\eta$ that perfectly maintains the zeroth-order topological information of $G_k$, it holds that $\eta_{\tau} \geq \eta$.

Furthermore,
\begin{itemize}
    \item When $G_k$ is a fully connected~\footnote{Also referred to as ``dense" in some following results.} layer
    $$\eta_{\tau} = \frac{m_k \cdot n_k}{m_k + n_k -1 }$$
    \item When $G_k$ is a recurrent layer with $\ell_k$ hidden units
    $$\eta_{\tau} = \frac{\ell_k^2}{2\ell_k-1 }$$
    \item When $G_k$ is a convolutional layer
    $$\eta_{\tau} = \frac{n_k \cdot f_1 \cdot f_2}{m_k + n_k - 1}$$
    with $(f_1,f_2)$ being size of the convolutional kernel.
    $m_k, n_k$ are the input and output sizes respectively of the spatial activations.
\end{itemize}
\begin{proof} 
In Appendix \ref{appx/proofs}. 
\end{proof}
\end{theorem}

\subsection{Bounds on the MST -- MP Fraction of Overlap}

We now state a lower bound on the expected overlap in the MST of a layer $G_k$ with its top-$\alpha$ weights, where $\alpha$ is the number of weights in its MST, to get a sense of how much of the zeroth-order topological information in a layer might be retained if we prune it down to its topologically critical compression ratio simply using the magnitude, thereby formally quantifying IMP's efficacy.

\begin{theorem}
\label{thm:overlap}
For a fully connected layer $G_k$ with normalized weights $\mathcal{W}^{\prime}_k$ joining $m_k$ nodes at the input to $n_k$ nodes at the output, for a compression ratio of $\eta_{\tau}$, the fraction of overlap expected in its top-$\alpha$ weights by magnitude and those in its MST can be lower bounded as
$$
\mathbb{E}[X] \geq \frac{1}{m_k + n_k - 1} \cdot \sum^{j}_{i=0}\left(\frac{(m_k - i)(n_k - i)}{m_k \cdot n_k -i}\right)
$$
where $j = \operatorname{min}(m_k,n_k) \geq 2$ and X is the fraction of overlap between the two quantities of interest.
\vskip -0.2in
$$\textrm{If }j = 1, \mathbb{E}[X] = 1$$
\end{theorem}

\begin{proof}
In Appendix \ref{appx/proofs}.
\end{proof}

\begin{corollary}
If the fully connected graph $G_k$ is $p$-sparse, i.e, has a fraction of $p$ non-zero weights $\geq \alpha$,

$$
\mathbb{E}[X] \geq \operatorname{min}\left\{1,~ \frac{1}{m_k + n_k - 1} \cdot \sum^{j}_{i=0}\left(\frac{(m_k - i)(n_k - i)}{p \cdot m_k \cdot n_k -i}\right)\right\}
$$
where $j = \operatorname{min}(m_k,n_k) \geq 2$ and X is the fraction of overlap between the two quantities of interest.
\vskip -0.2in
$$\textrm{If }j = 1, \mathbb{E}[X] = 1$$
\end{corollary}

\begin{proof}
In Appendix \ref{appx/proofs}.
\end{proof}

We note here that these bounds only give us a sense for how much the topological complexity could be maintained;
The exact values for the same would rely on the caluculation of the neural persistence and therefore depend on the exact distribution of weights $\mathcal{W}_k^{\prime}$.

\subsection{Topological Iterative Magnitude Pruning}
\label{topologicalIMP}

The aforementioned insights and results thus naturally suggest a simple modification to the IMP algorithm that would ensure preservation of zeroth-order topological information in every layer.
Following a similar structure as the IMP algorithm in Section \ref{sec:imp_bg}, we now have Topological-IMP (T-IMP) that takes the following steps:

\begin{enumerate}
    \item Find the weights which form the MST and retain them, accounting for $\alpha$ weights out of $\frac{p}{N}$\% that one wishes to keep.
    \item From the remaining $\frac{p}{N}$\% - $\alpha$ weights to be retained at that iteration, pick those with the highest magnitudes.
    \item Retrain the network.
    \item Repreat the process $N$ times until the target sparsity-accuracy is reached.
\end{enumerate}

\section{Empirical Simulations \& Results}

\subsection{Topologically Critical Compression}

To see the practical significance of the topologically critical compression ratio and 
quantify the extent of pruning it can achieve, we experimented with  combinations of popular datasets and architectures (Table \ref{tab/eta_c_all}).
Additional experiments with MNIST, as well as compression details on a per layer basis are available in the appendix (Appendices \ref{appx/mnist_ratios}, \ref{appx/layers}).
Our overall insights from the investigations are as follows:
\begin{itemize}
    \item Fully connected layers are a lot more redundant and ergo compressible than convolutional layers, and this seems to hold true across dataset-architecture pairings.
    The compressability of these dense layers is what often seems to present incredibly high numbers for how compressable a particular model is.
    \item Amongst the convolutional architectures, inherently more efficient architectures (e.g., ResNet) are slightly less compressable than their more redundant counterparts such as the VGG, even by topological metrics.
\end{itemize}

\begin{table}[t]
\caption{Topologically Critical Compression: VGG11 \& ResNet}
\label{tab/eta_c_all}
\vskip 0.15in
\begin{center}
\begin{small}
\begin{sc}
\begin{tabular}{lll}
\toprule
\hfill & VGG11 ($\eta_{\tau}$) & ResNet ($\eta_{\tau}$) \\
\midrule
\hfill        & Conv: 4.2295  & Conv: 4.1508\\
CIFAR10$^{*}$ & Dense: 9.8273 & Dense: 8.7671\\
\hfill        & Final: 4.3914 & Final: 4.1679\\
\abovespace
\hfill          & Conv: 4.2295    & Conv: 4.1508\\
CIFAR100$^{*}$  & Dense: 83.7971  & Dense: 39.2638\\
\hfill          & Final: 6.9239   & Final: 4.4389\\
\abovespace
\hfill                     & Conv: 4.2672     & Conv: 4.3142\\
Tiny-ImageNet$^{\dagger}$  & Dense: 528.3911  & Dense: 144.0225\\
\hfill                     & Final: 183.3624  & Final: 6.13182\\
 
\bottomrule
\abovespace
$^{*}$ ResNet-20 \\
$^{\dagger}$ ResNet-18 \\
\end{tabular}
\end{sc}
\end{small}
\end{center}
\vskip -0.1in
\end{table}

\begin{figure}[ht]
\vskip 0.2in
\begin{center}
\centerline{\includegraphics[width=\columnwidth]{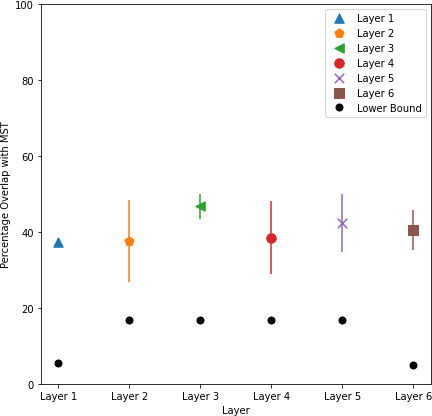}}
\caption{Top row: Mean percentage overlap (with standard deviations) that the top-$\alpha$ weights have with the MST computed over different layers of a fully-connected network trained on the MNIST dataset.
Black dots at the bottom represent the derived theoretical lower bound for the overlap for each of the respective layers.}
\label{fig:mst_overlap}
\end{center}
\vskip -0.2in
\end{figure}

\subsection{Bounds on the MST -- MP Fraction of Overlap}

The bound presented in Thm. \ref{thm:overlap} was also checked empirically with multiple simulations on different layers of the MNIST fully connected model (Fig.~\ref{fig:mst_overlap}).
While not the tightest, the bound (and simulations) still provided substantial support that MP does encourage preservation of zeroth-order topological features\footnote{See Appendix \ref{appx/random_probs} for a related discussion.} in the DNN weight space.
However, preservation of zeroth-order topology only forms part of the story concerning which weights ought to be kept when pruning DNNs.
Extensions to higher order homologies might perhaps help explain these discrepancies better. 


\section{Discussion \& Future Work}
\label{discussion}

In this work we presented a novel perspective on IMP leveraging a zeroth-order topological measure, viz., neural persistence.
The resulting insights now provide us with the opportunity to pursue some exciting avenues on both, theoretical and empirical fronts.

These include possible extensions of the stated bounds and subsequent theory to stratified graphs to explain global pruning, as well the extension of NP itself to beyond zeroth-order homology to potentially uncover a fuller picture of the topological complexities of DNNs.

Additionally, topological perspectives on LTH, weight rewinding, single-shot pruning, and other interesting IMP-adjacent phenomena could lead to not only insights into the interplay between DNN training and inference dynamics, but also topologically-motivated algorithms to achieve data- and compute-efficient deep learning pipelines. 



\section*{Acknowledgements}
The authors thank Georgia Tech's Graduate Student Association and College of Engineering for providing financial support to present this work, and the reviewers for their constructive feedback.
AB would also like to thank Nischita Kaza for helpful comments regarding the paper's exposition.

\clearpage
\bibliography{example_paper}
\bibliographystyle{icml2022}

\newpage
\appendix
\onecolumn

\section{Proofs}
\label{appx/proofs}

\begin{theorem}
Given a layer $G_k$ with weights $\mathcal{W}_k^{\prime}$ joining $m_k$ input nodes to $n_k$ output nodes, for any compression ratio $\eta$ that perfectly maintains the zeroth-order topological information of $G_k$ it holds that $\eta_{\tau} \geq \eta$.

Furthermore,
\begin{itemize}
    \item When $G_k$ is a fully connected (i.e., dense) layer
    $$\eta_{\tau} = \frac{m_k \cdot n_k}{m_k + n_k -1 }$$
    \item When $G_k$ is a recurrent layer with $\ell_k$ hidden units
    $$\eta_{\tau} = \frac{\ell_k^2}{2\ell_k-1 }$$
    \item When $G_k$ is a convolutional layer
    $$\eta_{\tau} = \frac{n_k \cdot f_1 \cdot f_2}{m_k + n_k - 1}$$
    with $(f_1,f_2)$ being size of the convolutional kernel.
    $m_k, n_k$ are the input and output sizes respectively of the spatial activations.
\end{itemize}
\begin{proof}

Any compression ratio $\eta$ that perfectly preserves the zeroth order topological information of the graph would need to preserve its entire MST (and therefore at least as many weights as the MST), thereby making $\eta \leq \eta_{\tau}$

For a bipartite graph having ${m_k, n_k}$ as its sets of disjoint vertices, $|\mathcal{W}_k^{\prime}| = m_k \cdot n_k$ and the MST has precisely $m_k + n_k - 1$ edges.
The expression for $\eta_{\tau}$ in the case of a fully connected layer then follows trivially from the previous facts combined with the definition of $\eta_{\tau}$

Likewise, in the case of a recurrent layer, $m_k = n_k = \ell_k$ and the number of parameters to be stored at anytime would be exactly the same as that of a fully connected layer of the same dimensions, since the exact same weights are shared across all the unrolled instances of the recurrent layer and therefore the compression ratio remains equal across all of them.

In the case of convolutional layers however, we first need to think of the process of convolution with a kernel of dimensions $(f_1,f_2)$ as the matrix multiplication of the vectorized input (i.e., activations of the preceding layer) with an appropriately-sized Toeplitz matrix having a total of $n_k \cdot f_1 \cdot f_2$ non-zero elements.
More precisely, we would have:
\begin{itemize}
    \item $m_k$ input nodes where $m_k := \prod_{i=1}^{2} 2pad_i+s_i$, with $s_i, pad_i$ being the input size and padding in the respective spatial directions.
    \item $n_k$ output nodes where $n_k := \prod_{i=1}^{2}\lfloor{\frac{s_i+2pad_1-f_i}{t_i}}\rfloor$ and $f_i, s_i, t_i, pad_i$ being the size of the kernel, input, padding, and stride in the appropriate spatial direction.
    \item A sparse, fully-connected layer whose weights when represented as a matrix of dimensions $(m_k,n_k)$ follow a Toeplitz structure, with the same $f_1 \cdot f_2$ weights being cyclically shifted in every column of the matrix.
\end{itemize}
   
With $(m_k, n_k)$ nodes in the bipartite sense once again, the resulting $\eta_{\tau}$ follows that of a corresponding fully connected layer with the caveat that we have only $n_k \cdot f_1 \cdot f_2$ non-zero weights in the layer to begin with.

\end{proof}
\end{theorem}

\newpage
\begin{theorem}
\label{thm:overlap_appx}
For a fully connected layer $G_k$ with normalized weights $\mathcal{W}^{\prime}_k$ joining $m_k$ nodes at the input to $n_k$ nodes at the output, for a compression ratio of $\eta_{\tau}$, the fraction of overlap expected in its top-$\alpha$ weights by magnitude and those in its MST can be lower bounded as
$$
\mathbb{E}[X] \geq \frac{1}{m_k + n_k - 1} \cdot \sum^{j}_{i=0}\left(\frac{(m_k - i)(n_k - i)}{m_k \cdot n_k -i}\right)
$$
where $j = \operatorname{min}(m_k,n_k) \geq 2$ and X is the fraction of overlap between the two quantities of interest.
\vskip -0.2in
$$\textrm{If }j = 1, \mathbb{E}[X] = 1$$
\end{theorem}

\begin{proof}
The proof for the above statement relies on being able to lower bound individually the probabilities of the graph's top-$\alpha$ weights by magnitude being in its MST.
In order for this to be the case, every new weight added must never form a cycle, i.e., either join two nodes both of which were disconnected from all other vertices in the graph before, or join one new node to the some other connected component in the graph.
We only consider instances where the former occurs.

Starting with the highest weight (i.e., $w_0^{\prime}$ which is normalized to 1, and going in decreasing order of normalized magnitude), the probability of the weight $w_i$ connecting two previously \emph{isolated} vertices in the bipartite graph $G_k$ is equal to the quantity $\frac{1}{m_k + n_k -1} \cdot \left(\frac{(m_k - i)(n_k - i)}{m_k \cdot n_k -1}\right)$, since
$m_k + n_k - 1$ is the cardinality of the set of top-$\alpha$ weights, and
$\frac{(m_k - i)(n_k - i)}{m_k \cdot n_k - 1}$ is the minimum fraction of the number of possible isolated vertices to total number of edges.

We subsequently sum the probabilities over until we reach the last $w_i$ such that $i = j$ (which by definition is $\geq 2$), since beyond that, the minimum possible number of isolated vertices is no longer definitely $ > 0$.

If $j = 1$, the MST overlaps completely with the entire set of weights for $G_k$, making $\mathbb{E}[X] = 1$.

In both instances, we assume the bipartite graph $G_k$ to be complete.
\end{proof}

\begin{corollary}
If the fully connected graph $G_k$ is $p$-sparse, i.e, has a fraction of $p$ non-zero weights $\geq \alpha$,

$$
\mathbb{E}[X] \geq \operatorname{min}\left\{1,~ \frac{1}{m_k + n_k - 1} \cdot \sum^{j}_{i=0}\left(\frac{(m_k - i)(n_k - i)}{p \cdot m_k \cdot n_k -i}\right)\right\}
$$
where $j = \operatorname{min}(m_k,n_k) \geq 2$ and X is the fraction of overlap between the two quantities of interest.
\vskip -0.2in
$$\textrm{If }j = 1, \mathbb{E}[X] = 1$$
\end{corollary}

\begin{proof}
The proof for the corollary follows from the proof of Thm. \ref{thm:overlap_appx}, with the simple modification that if the graph is sparser by a multiplicative factor $p$, the total number of weights in the graph now becomes $p\cdot m_k\cdot n_k$.

Note that we make the slight assumption that all the nodes of the graph are still connected, despite the sparsity.
\end{proof}


\clearpage
\section{Super-level Set Filtration}
\label{appx/superlevel_filtration}
\begin{figure}[ht]
\begin{center}
\centerline{\includegraphics[width=\columnwidth]{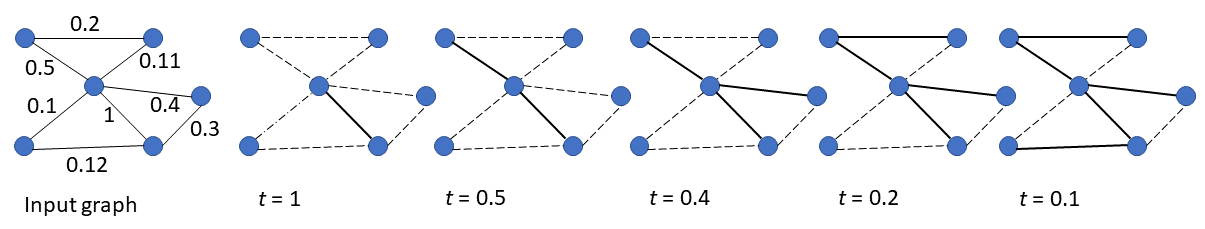}}
\caption{Example of a super-level set filtration on the edges of a weighted graph.
Since the filtration is being carried out to construct the zeroth-order persistence diagram, we start with all nodes as individual connected components, and edges are only taken into consideration when their weight is equal to the threshold, subject to the condition they do not create cycles (since that would no longer be changing the number of connected components in consideration).
The filtration finally ends up with a Maximum Spanning Tree (MST) of the graph.
The diagram above is adapted from Figure 6 in the appendix of \citeauthor{lacombe2021topological}}
\label{fig/superlevel_filtration}
\end{center}
\vskip -0.2in
\end{figure}

\section{Random Probabilities of MST -- Top-$\alpha$ Weights Overlapping}
\label{appx/random_probs}

At first glance, our results quantifying the overlap between the top-$\alpha$ weights and the MST of various layers of a trained DNN (Fig. \ref{fig:mst_overlap}) seem modest, both empirically (mean $\sim 40\%$) and theoretically (minimum lower bound $\sim 5\%$). However, it is important to keep in mind the random probabilities of these events occurring;
In particular, for a layer $G_k$ that is represented as a complete bipartite graph with $m_k$ input and $n_k$ output nodes, the probability of two random subsets of $\alpha$ weights each having exactly $w$ of them overlapping is given as
$$p_{\alpha, w}(overlap) = \binom{\alpha}{w}\left(\frac{\alpha}{m_k \cdot n_k}\right)^{w}\left(\frac{m_k \cdot n_k-\alpha}{m_k \cdot n_k}\right)^{\alpha-w} 
$$
while the probability of the two random sets of size $\alpha$ each having \emph{at least} $w$ weights overlapping is
$$p_{\alpha, w_{++}}(overlap) = \sum_{i=w}^{\alpha} \binom{\alpha}{i}\left(\frac{\alpha}{m_k \cdot n_k}\right)^{i}\left(\frac{m_k \cdot n_k-\alpha}{m_k \cdot n_k}\right)^{\alpha-i}$$

This implies, for $\alpha=m_k+n_k-1$ and $(m_k,n_k) = (784,100), (100,100), (100,10)$ the probabilities of having:
\begin{itemize}
    \item Exactly 5\% overlap = $7.4 \times 10^{-6},~0.012,~0.011$
    \item Exactly 40\% overlap = $\sim 0,~ 2.40 \times 10^{-79},~8.79 \times 10^{-15}$
    \item At least 5\% overlap = $9.4 \times 10^{-6},~0.019,~0.994$
    \item At least 40\% overlap = $\sim 0,~2.48 \times 10^{-79},~1.07 \times 10^{-14}$
\end{itemize}

As we can see, the $\sim 40\%$ overlap that we see empirically across layers between their MST and top-$\alpha$ weights is actually quite significant, and therefore a good indicator that our hypothesis regarding magnitude pruning encouraging zeroth-order topological feature preservation might indeed be true.
Even the much more modest theoretical bounds are fairly informative, except perhaps when dealing with the last layer, with a very low expected fraction of overlap.

\clearpage
\section{MNIST Architecture Details}
\label{appx/MNIST_archs}
The layerwise architectural details for the fully connected (FCN) and convolutional (CNN) architectures used with the MNIST dataset are provided in Tables~\ref{tab/MNIST_Dense_arch} and~\ref{tab/MNIST_Conv_arch} respectively.
They are the same as those implemented in the SynFlow \cite{tanaka2020pruning} GitHub repository located at \url{https://github.com/ganguli-lab/Synaptic-Flow}.

\begin{table} [H]
\caption{MNIST (FCN)}
\label{tab/MNIST_Dense_arch}
\vskip 0.15in
\begin{center}
\begin{small}
\begin{sc}
\begin{tabular}{lc}
\toprule
Layer & Details \\
\midrule
Dense Layer 1 & Input Dim: 784, Output Dim: 100 \\
Dense Layer 2 & Input Dim: 100, Output Dim: 100 \\
Dense Layer 3 & Input Dim: 100, Output Dim: 100 \\
Dense Layer 4 & Input Dim: 100, Output Dim: 100 \\
Dense Layer 5 & Input Dim: 100, Output Dim: 100 \\
Dense Layer 6 & Input Dim: 100, Output Dim: 10 \\
\bottomrule
\end{tabular}
\end{sc}
\end{small}
\end{center}
\vskip -0.1in
\end{table}

\begin{table}[H]
\caption{MNIST (CNN)}
\label{tab/MNIST_Conv_arch}
\vskip 0.15in
\begin{center}
\begin{small}
\begin{sc}
\begin{tabular}{lc}
\toprule
Layer & Details \\
\midrule
Convolutional Layer 1 & Filters: 32, Kernel = 3x3, Padding = 1\\
Convolutional Layer 2 & Filters: 32, Kernel = 3x3, Padding = 1\\
Dense Layer & Input Dim: 25,088, Output Dim: 10\\
\bottomrule
\end{tabular}
\end{sc}
\end{small}
\end{center}
\vskip -0.1in
\end{table}

\section{MNIST Topologically Critical Compression Ratios}
\label{appx/mnist_ratios}
\begin{table}[ht]
\caption{Topologically Critical Compression: MNIST (FCN)}
\label{tab/MNIST_Dense}
\vskip 0.15in
\begin{center}
\begin{small}
\begin{sc}
\begin{tabular}{lc}
\toprule
Layer & Compression Ratio ($\eta_{\tau}$) \\
\midrule
Dense Layer 1  & 88.78822 \\
Dense Layer 2  & 50.25126 \\
Dense Layer 3  & 50.25126 \\
Dense Layer 4  & 50.25126 \\
Dense Layer 5  & 50.25126 \\
Dense Layer 6  & 9.17431  \\
\abovespace
Final Compression & 66.77852 \\
\bottomrule
\end{tabular}
\end{sc}
\end{small}
\end{center}
\vskip -0.1in
\end{table}

\begin{table} [ht]
\caption{Topologically Critical Compression: MNIST (CNN)}
\label{tab/MNIST_Conv}
\vskip 0.15in
\begin{center}
\begin{small}
\begin{sc}
\begin{tabular}{lc}
\toprule
Layer & Compression Ratio ($\eta_{\tau}$) \\
\midrule
Conv Layer 1 & 4.19251 \\
Conv Layer 2 & 4.19251 \\
Dense Layer           & 9.99641 \\
\abovespace
Final Compression & 9.31005 \\
\bottomrule
\end{tabular}
\end{sc}
\end{small}
\end{center}
\vskip -0.1in
\end{table}

\clearpage
\section{Layerwise Compression Ratios for Different Architecture + Dataset Pairings}
\label{appx/layers}

\begin{table} [H]
\caption{Topological Critical Compression: VGG11 + CIFAR10}
\label{tab/VGG11&CIFAR10}
\vskip 0.15in
\begin{center}
\begin{small}
\begin{sc}
\begin{tabular}{lc}
\toprule
Layer & Compression Ratio ($\eta_{\tau}$) \\
\midrule
Convolutional Layer 1 & 4.22946 \\
Convolutional Layer 2 & 4.22946 \\
Convolutional Layer 3 & 4.22946 \\
Convolutional Layer 4 & 4.22946 \\
Convolutional Layer 5 & 4.22946 \\
Convolutional Layer 6 & 4.22946 \\
Convolutional Layer 7 & 4.22946 \\
Convolutional Layer 8 & 4.22946 \\
Convolutional Layer 9 & 4.22946 \\
Dense Layer & 9.82726 \\
\abovespace
Final Compression & 4.39139  \\
\bottomrule
\end{tabular}
\end{sc}
\end{small}
\end{center}
\vskip -0.1in
\end{table}

\begin{table} [H]
\caption{Topological Critical Compression: VGG11 + CIFAR100}
\label{tab/VGG11&CIFAR100}
\vskip 0.15in
\begin{center}
\begin{small}
\begin{sc}
\begin{tabular}{lc}
\toprule
Layer & Compression Ratio ($\eta_{\tau}$) \\
\midrule
Convolutional Layer 1 & 4.22946 \\
Convolutional Layer 2 & 4.22946 \\
Convolutional Layer 3 & 4.22946 \\
Convolutional Layer 4 & 4.22946 \\
Convolutional Layer 5 & 4.22946 \\
Convolutional Layer 6 & 4.22946 \\
Convolutional Layer 7 & 4.22946 \\
Convolutional Layer 8 & 4.22946 \\
Convolutional Layer 9 & 4.22946 \\
Dense Layer & 83.79705 \\
\abovespace
Final Compression & 6.9239 \\
\bottomrule
\end{tabular}
\end{sc}
\end{small}
\end{center}
\vskip -0.1in
\end{table}

\begin{table} [H]
\caption{Topological Critical Compression: VGG11 + Tiny ImageNet}
\label{tab/VGG11&TinyImageNet}
\vskip 0.15in
\begin{center}
\begin{small}
\begin{sc}
\begin{tabular}{lc}
\toprule
Layer & Compression Ratio ($\eta_{\tau}$) \\
\midrule
Convolutional Layer 1 & 4.36209 \\
Convolutional Layer 2 & 4.22946 \\
Convolutional Layer 3 & 3.97927 \\
Convolutional Layer 4 & 3.97927 \\
Convolutional Layer 5 & 3.53374 \\
Convolutional Layer 6 & 3.53374 \\
Convolutional Layer 7 & 2.82353 \\
Convolutional Layer 8 & 2.82353 \\
Dense Layer 1 & 682.88896 \\
Dense Layer 2 & 512.25012 \\
Dense Layer 3 & 167.45707 \\
\abovespace
Final Compression & 183.3624 \\
\bottomrule
\end{tabular}
\end{sc}
\end{small}
\end{center}
\vskip -0.1in
\end{table}

\begin{table} [H]
\caption{Topological Critical Compression: ResNet-20 + CIFAR10}
\label{tab/Res20&CIFAR10}
\vskip 0.15in
\begin{center}
\begin{small}
\begin{sc}
\begin{tabular}{lc}
\toprule
Layer & Compression Ratio ($\eta_{\tau}$) \\
\midrule
Convolutional Layer 1 & 4.22946 \\
Convolutional Layer 2 & 4.22946 \\
Convolutional Layer 3 & 4.22946 \\
Convolutional Layer 4 & 4.22946 \\
Convolutional Layer 5 & 4.22946 \\
Convolutional Layer 6 & 4.22946 \\
Convolutional Layer 7 & 4.22946 \\
Convolutional Layer 8 & 3.97927 \\
Convolutional Layer 9 & 3.97927 \\
Convolutional Layer 10 & 3.97927 \\
Convolutional Layer 11 & 3.97927 \\
Convolutional Layer 12 & 3.97927 \\
Convolutional Layer 13 & 3.97927 \\
Convolutional Layer 14 & 3.53374 \\
Convolutional Layer 15 & 3.53374 \\
Convolutional Layer 16 & 3.53374 \\
Convolutional Layer 17 & 3.53374 \\
Convolutional Layer 18 & 3.53374 \\
Convolutional Layer 19 & 3.53374 \\
Dense Layer & 8.76712 \\
\abovespace
Final Compression & 4.16786 \\
\bottomrule
\end{tabular}
\end{sc}
\end{small}
\end{center}
\vskip -0.1in
\end{table}

\begin{table} [H]
\caption{Topological Critical Compression: ResNet-20 + CIFAR100}
\label{tab/Res20&CIFAR100}
\vskip 0.15in
\begin{center}
\begin{small}
\begin{sc}
\begin{tabular}{lc}
\toprule
Layer & Compression Ratio ($\eta_{\tau}$) \\
\midrule
Convolutional Layer 1 & 4.22946 \\
Convolutional Layer 2 & 4.22946 \\
Convolutional Layer 3 & 4.22946 \\
Convolutional Layer 4 & 4.22946 \\
Convolutional Layer 5 & 4.22946 \\
Convolutional Layer 6 & 4.22946 \\
Convolutional Layer 7 & 4.22946 \\
Convolutional Layer 8 & 3.97927 \\
Convolutional Layer 9 & 3.97927 \\
Convolutional Layer 10 & 3.97927 \\
Convolutional Layer 11 & 3.97927 \\
Convolutional Layer 12 & 3.97927 \\
Convolutional Layer 13 & 3.97927 \\
Convolutional Layer 14 & 3.53374 \\
Convolutional Layer 15 & 3.53374 \\
Convolutional Layer 16 & 3.53374 \\
Convolutional Layer 17 & 3.53374 \\
Convolutional Layer 18 & 3.53374 \\
Convolutional Layer 19 & 3.53374 \\
Dense Layer & 39.2638 \\
\abovespace
Final Compression & 4.4389 \\
\bottomrule
\end{tabular}
\end{sc}
\end{small}
\end{center}
\vskip -0.1in
\end{table}

\begin{table} [H]
\caption{Topological Critical Compression: ResNet-20 + Tiny ImageNet}
\label{tab/Res18&TinyImageNet}
\vskip 0.15in
\begin{center}
\begin{small}
\begin{sc}
\begin{tabular}{lc}
\toprule
Layer & Compression Ratio ($\eta_{\tau}$) \\
\midrule
Convolutional Layer 1 & 4.36209 \\
Convolutional Layer 2 & 4.36209 \\
Convolutional Layer 3 & 4.36209 \\
Convolutional Layer 4 & 4.36209 \\
Convolutional Layer 5 & 4.36209 \\
Convolutional Layer 6 & 4.22946 \\
Convolutional Layer 8 & 4.22946 \\
Convolutional Layer 9 & 4.22946 \\
Convolutional Layer 10 & 3.97927 \\
Convolutional Layer 11 & 3.97927 \\
Convolutional Layer 12 & 3.97927 \\
Convolutional Layer 13 & 3.97927 \\
Convolutional Layer 14 & 3.53374 \\
Convolutional Layer 15 & 3.53374 \\
Convolutional Layer 16 & 3.53374 \\
Convolutional Layer 17 & 3.53374 \\
Dense Layer & 144.0225 \\
\abovespace
Final Compression & 6.13182 \\
\bottomrule
\end{tabular}
\end{sc}
\end{small}
\end{center}
\vskip -0.1in
\end{table}

\section{Architecture + Dataset Pairings}
\label{appx/arch_dataset_pairings}

\begin{table}[ht]
\caption{Architectures and Dataset Pairings}
\label{tab/experimental_setup}
\vskip 0.15in
\begin{center}
\begin{small}
\begin{sc}
\begin{tabular}{lc}
\toprule
Architecture & Dataset \\
\midrule
Fully Connected MNIST & MNIST\\
Convolutional MNIST & MNIST\\
\abovespace
\hfill & CIFAR10  \\ 
VGG11  & CIFAR100 \\
\hfill & Tiny-ImageNet  \\
\abovespace
\hfill & CIFAR10$^{*}$  \\ 
ResNet & CIFAR100$^{*}$ \\
\hfill & Tiny-ImageNet$^{\dagger}$  \\
\bottomrule
\abovespace
$^{*}$ ResNet-20 \\
$^{\dagger}$ ResNet-18 \\
\end{tabular}
\end{sc}
\end{small}
\end{center}
\vskip -0.1in
\end{table}

\clearpage
\section{Normalized Weight Value Comparisons Between MST Weights and Top-$\alpha$ Weights}
\label{appx/mst_overlap}
\begin{figure}[ht]
\vskip 0.2in
\begin{center}
\centerline{\includegraphics[width=\columnwidth]{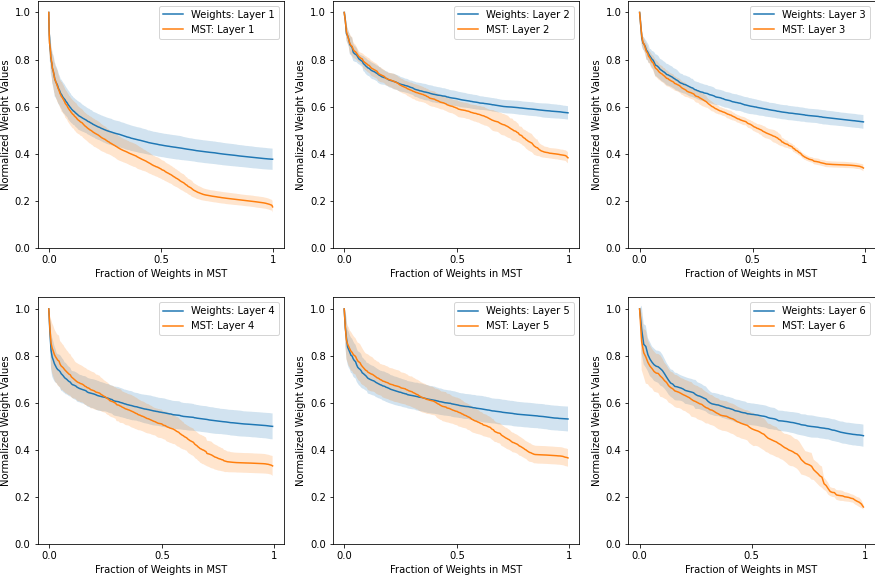}}
\caption{Visualization of the normalized weight values in the top-$\alpha$ weights and the MST weights for different layers in the trained, 6-layer MNIST FCN over 5 trials.}
\label{fig/MST_6}
\end{center}
\vskip -0.2in
\end{figure}

\end{document}